\documentclass[10pt]{article}
\usepackage{amsmath}
\usepackage{booktabs}
\usepackage{fancyhdr}
\usepackage[dvipsnames]{xcolor}
\usepackage[most]{tcolorbox}

\usepackage[round]{natbib}
\usepackage{geometry}
\usepackage{graphicx}

\usepackage{charter}
\usepackage{hyperref}
\hypersetup{colorlinks=true,allcolors=BlueViolet}

\title{\textsc{\textbf{Transformers vs. Recurrent Models for Estimating
Forest Gross Primary Production}}}

\author{
David Montero\textsuperscript{1,2,*}\quad 
Miguel D. Mahecha\textsuperscript{1,2}\quad
Francesco Martinuzzi\textsuperscript{3}\\
César Aybar\textsuperscript{4}\quad 
Anne Klosterhalfen\textsuperscript{5}\quad
Alexander Knohl\textsuperscript{5}\\
Jesús Anaya\textsuperscript{6}\quad
Clemens Mosig\textsuperscript{1}\quad
Sebastian Wieneke\textsuperscript{1} \\
\small\textsuperscript{1}IEF, Leipzig University\quad
\textsuperscript{2}iDiv\quad
\textsuperscript{3}MPI PKS\quad
\textsuperscript{4}IPL, Universitat de Val\`encia\\
\small\textsuperscript{5}Bioclimatology, University of Göttingen\quad
\textsuperscript{6}GEMA, Universidad de Medellín\\
\small*Corresponding author: \texttt{david.montero@uni-leipzig.de}
}
\date{}

\begin{document}

\maketitle

\begin{abstract}
Monitoring the spatiotemporal dynamics of forest CO$_2$ uptake (Gross Primary Production, GPP), remains a central challenge in terrestrial ecosystem research.
While Eddy Covariance (EC) towers provide high-frequency estimates, their limited spatial coverage constrains large-scale assessments. Remote sensing offers a scalable alternative, yet most approaches rely on single-sensor spectral indices and statistical models that are often unable to capture the complex temporal dynamics of GPP.
Recent advances in deep learning (DL) and data fusion offer new opportunities to better represent the temporal dynamics of vegetation processes, but comparative evaluations of state-of-the-art DL models for multimodal GPP prediction remain scarce.
Here, we explore the performance of two representative models for predicting GPP: 1) GPT-2, a transformer architecture, and 2) Long Short-Term Memory (LSTM), a recurrent neural network, using multivariate inputs.
Overall, both achieve similar accuracy. But, while LSTM performs better overall, GPT-2 excels during extreme events. Analysis of temporal context length further reveals that LSTM attains similar accuracy using substantially shorter input windows than GPT-2, highlighting an accuracy-efficiency trade-off between the two architectures. Feature importance analysis reveals radiation as the dominant predictor, followed by Sentinel-2, MODIS land surface temperature, and Sentinel-1 contributions.
Our results demonstrate how model architecture, context length, and multimodal inputs jointly determine performance in GPP prediction, guiding future developments of DL frameworks for monitoring terrestrial carbon dynamics.
\end{abstract}

\thispagestyle{fancy}
\renewcommand{\headrulewidth}{0pt}
\renewcommand{\footrulewidth}{0pt}
\fancyhf{}
\cfoot{\scriptsize{© 2025 IEEE. Personal use of this material is permitted. Permission
from IEEE must be obtained for all other uses, in any current or future
media, including reprinting/republishing this material for advertising or
promotional purposes, creating new collective works, for resale or
redistribution to servers or lists, or reuse of any copyrighted
component of this work in other works.}}

\section{Introduction}


Gross Primary Production (GPP) is a key indicator of ecosystem functioning and represents the largest carbon flux from the atmosphere to terrestrial ecosystems \citep{friedlingstein2025carbon_budget_2024}. Accurate GPP quantification is essential for assessing the role of forests in the global carbon cycle and their capacity to offset anthropogenic CO$_2$ emissions \citep{bonan2008forests_climate_change}. Forests act as major carbon sinks, providing a critical negative feedback to climate change \citep{pan2011large_persistent_carbon_sink}. However, extreme events such as droughts and heatwaves can drastically suppress GPP through physiological stress responses at physiological and molecular levels \citep{krich2022decoupling,sato2024complex}. These reductions may temporarily turn forests from carbon sinks into sources, amplifying climate-carbon feedbacks \citep{sippel2018drought}. Understanding and predicting GPP dynamics under both normal and extreme conditions is therefore essential for monitoring forest productivity and anticipating the impacts of climate extremes on terrestrial carbon balance \citep{zscheischler2014carbon, bastos2023joint}.


Traditional estimates of GPP rely on in-situ Eddy Covariance (EC) measurements (e.g., Ameriflux, ChinaFLUX, Fluxnet, ICOS), which provide accurate flux estimates but have limited spatial coverage \citep{kumar2016understanding,mahecha2017detecting}. Remote sensing offers a scalable alternative, typically using optical data to derive Vegetation Indices (VIs). The structural and pigment information captured by these VIs often correlates with photosynthetic activity, allowing them to serve as proxies for GPP across vegetation types \citep{campsvalls2021kndvi}. Combining multiple VIs, particularly those exploiting red-edge information, has further improved GPP prediction accuracy \citep{pabonmoreno2022potential_s2_gpp}. However, VIs primarily capture canopy greenness and structure, showing limited sensitivity to physiological stress. During droughts or heatwaves, photosynthesis declines rapidly due to stomatal regulation and the activation of reactive oxygen species \citep{sato2024complex}, while greenness remains largely unchanged \citep{vicca2016remotely_sensed_drought,hoekvandijke2023comparing_forests_grasslands_drought}, leading to underestimation of GPP reductions during extreme events.


Because optical VIs poorly capture physiological stress \citep{vicca2016remotely_sensed_drought,pabonmoreno2022potential_s2_gpp,hoekvandijke2023comparing_forests_grasslands_drought}, complementary sensing modalities such as Land Surface Temperature (LST) and Synthetic Aperture Radar (SAR) may provide valuable information for improving GPP estimation, especially during climate extremes \citep{hoekvandijke2023comparing_forests_grasslands_drought,konings2021forest_drought_vwc_radar}. LST, derived from thermal infrared observations, reflects surface energy balance and vegetation stress: reduced stomatal conductance limits evaporative cooling, leading to higher canopy temperatures \citep{miralles2018land_atmosphere_droughts_heatwaves,hoekvandijke2023comparing_forests_grasslands_drought}. SAR, in turn, is sensitive to canopy structure and water content; droughts and heatwaves reduce vegetation water content and diminish canopy structural complexity, leading to lower backscatter \citep{konings2021forest_drought_vwc_radar}. Its cloud-penetrating capability further enables consistent monitoring when optical data are unavailable \citep{chen2024gpp_microwave_biomes}. The joint use of optical, thermal, and radar data thus enhances the ability to capture both structural and physiological responses driving GPP variability.


Beyond multimodal observations, accurate GPP prediction requires capturing both the temporal evolution of ecosystems and their geographical context. Historical vegetation states are particularly relevant under climate extremes, where stress-induced degradation can reduce resilience \citep{anderegg2015tree_mortality_changing_climate}. Accurately representing these legacy effects requires models that account for temporal dependencies in the data. Traditional machine learning models represent temporal dynamics through engineered features \citep{kmrinen2023spatiotemporal_lag_nee}, whereas Deep Learning (DL) architectures can inherently learn complex temporal dependencies and nonlinear interactions among multimodal inputs \citep{camps2025artificial}. Recurrent neural networks such as Long Short-Term Memory (LSTM) models \citep{hochreiter1997lstm} have been used for GPP prediction \citep{montero2024rnns_gpp_igarss}, but they often struggle to retain information over long time periods \citep{martinuzzi2024learning}. Transformer architectures \citep{vaswani2017attention} overcome this limitation by directly attending to any point in the input sequence, enabling effective use of extended context windows for GPP estimation \citep{nakagawa2023gpp_tft}.
Geographical context is equally important. Geographic encodings (latitude, longitude) can distort spatial relationships \citep{montero2024avenues}, while sine-cosine or spherical harmonic representations provide smoother positional features \citep{russwurm2023spherical_harmonics}. Moreover, potential radiation offers a physically grounded proxy that jointly captures spatial and temporal dependencies by representing potentially available solar energy, an intrinsic driver of photosynthesis.


Despite the growing use of  DL for estimating forest GPP, direct comparisons between recur\-rence\--based models and Transformers remain scarce. This gap is particularly relevant during extreme weather events, when GPP departs from expected dynamics and model robustness becomes even more critical. Existing studies have largely focused on overall accuracy rather than evaluating how effectively different architectures retain memory of past vegetation states or capture anomalous GPP responses to stress \citep{martinuzzi2024learning}. Moreover, the relative importance of multimodal predictors and the role of physically based spatiotemporal encodings such as potential radiation have not been systematically assessed. 


This study addresses the above-mentioned gaps by comparing Transformer and LSTM models for estimating daily forest GPP from multimodal satellite time series. Our objectives are: 1) to evaluate their capacity to represent temporal memory, 2) to assess overall performance and under extreme conditions, and 3) to quantify modality contributions.

\section{Methods}

\subsection{Data Preparation}

The study covered 2016-2020, coinciding with Sentinel-2 availability and the ICOS 2020 Warm Winter dataset \citep{icos2022warmwinter}.

\subsubsection{Target Feature}
\label{sec:gpp}

Daily GPP data were obtained from ICOS (ONEflux pipeline, \texttt{GPP\_NT\_VUT\_REF}) \citep{pastorello2020oneflux_fluxnet2015}. Sites were filtered using CORINE 2018 Land Cover, retaining forest-dominated areas ($>$70\% cover within 1 km) and excluding urban or non-European locations. Timesteps with $<$70\% high-quality fluxes (\texttt{NEE\_VUT\_REF\_QC}) or negative GPP values were removed, and sites with $<$60\% valid data discarded. Nineteen sites remained, spanning temperate to boreal forests (12 ENF: CZ-RAJ, DE-RuW, DE-Tha, FI-Hyy, FI-Let, FI-Va, IT-Lav, IT-Ren, SE-Htm, SE-Nor, SE-Ros, SE-Svb, 4 DBF: CZ-Stn, DE-Hai, DE-HoH, FR-Fon, 3 MF: BE-Vie, CH-Lae, CZ-Lnz). Anomalies were defined as deviations from the mean seasonal cycle; the lowest and highest 10\% were flagged as negative (GPP$^-$) and positive (GPP$^+$) extremes. Only sequences of at least five consecutive flagged days were considered as GPP extremes.

\subsubsection{Input Features}
\label{sec:inputs}

Sentinel-2 L2A cubes (10 m) were obtained from the Planetary Computer STAC. Non-10 m bands were resampled, and cloud, shadow, and snow pixels masked using CloudSEN12 \citep{Aybar2022cloudsen12} and Fmask v3.2 \citep{zhu2015fmask}. Reflectances were converted to Nadir BRDF-Adjusted Reflectance using the c-factor method \citep{montero2024facilitating_sen2nbar}. Forest pixels were spatially aggregated by computing the median within 1 km buffers around the EC towers. To summarize canopy optical properties, we computed a comprehensive set of 122 Sentinel-2 VIs (ASI v0.5.0 \citep{montero2023asi}) and then reduced their strong collinearity via PCA; the first 18 components ($>$99\% variance) served as the S2 feature set. This approach stabilizes training with limited data, mitigates multicollinearity, and retains spectral sensitivities while avoiding index selection bias.

Sentinel-1 (SAR) RTC $\gamma^0$ backscatter data (VV and VH, ascending and descending orbits) were retrieved from the Planetary Computer STAC. The Dual-Polarized Radar Vegetation Index (DpRVI${\textrm{VV}}$) \citep{nasirzadehdizaji2019dprvi} was computed from $\gamma^0$ values in linear units, which were subsequently converted to decibels (dB). The five resulting features (DpRVI${\textrm{VV}}$; $\gamma^0_{VV}$, and $\gamma^0_{VH}$ in both linear and dB units) were spatially aggregated (median) within 50–1000 m buffers around each EC tower to form the S1 modality.

Daily MODIS LST v6.1 (Terra MOD11A1, Aqua MYD11A1) were processed via Google Earth Engine. Daytime and nighttime LST from both satellites were spatially smoothed (3$\times$3 mean kernel) and extracted at each EC tower, yielding four features that composed the LST modality.

\subsubsection{Spatiotemporal Encoding}
\label{sec:rad}

Clear-sky shortwave radiation (R$_{so}$, W m$^{-2}$) \citep{reda2008solar_position_algorithm} was calculated hourly at each EC tower location, and the daily mean was computed. R$_{so}$ encodes solar-geometry factors (day length, declination, zenith angle) and serves as a physically based proxy for photosynthetically available energy, treated as an independent modality.

\subsubsection{Tokenization and Feature Representation}
\label{sec:tokenization}

Each day was represented as a token $\mathbf{x}_t$, a 28-dimensional vector combining 18 S2, 5 S1, 4 LST features, and R$_{so}$. Missing values were linearly interpolated to produce complete sequences. The resulting daily series, with a look-back context window of $k=120$, were used as input for both LSTM and GPT-2 models.

\subsection{Model Architectures}

We used two deep learning architectures to model daily forest GPP from multimodal time-series inputs: a Long Short-Term Memory (LSTM) network and a Transformer model based on GPT-2. Both received identical input sequences (120 days $\times$ 28 features) and output a single daily GPP estimate.

\subsubsection{LSTM}

LSTM \citep{hochreiter1997lstm} processes sequential data through recurrent cells that maintain an internal sequential memory of past inputs, enabling the model to learn temporal dependencies while mitigating vanishing-gradient effects. The network consists of $L$ stacked layers whose final hidden state is projected through a linear layer to produce the daily GPP prediction.

\subsubsection{GPT-2}

GPT-2 \citep{radford2019language} employs multi-head self-attention \citep{vaswani2017attention} to compute contextualized representations of all time steps simultaneously, allowing direct access to long-range dependencies. Inputs are linearly projected into the model dimension and augmented with learned positional encodings. The resulting embeddings are passed through $L$ stacked Transformer decoder blocks, followed by a linear projection to predict daily GPP.

\subsection{Training Strategy}

\subsubsection{Hyperparameters Tuning}

Data were partitioned temporally: 2016-2018 for training, 2019 for testing, and 2020 for validation. Hyperparameters were optimized using HyperBand \citep{li2018hyperband}. For the LSTM, the hidden size and number of layers were tuned, whereas for GPT-2 we optimized the model dimension, feed-forward width, number of attention heads, and decoder layers. A dropout rate of 0.3 was applied to all layers.

\subsubsection{Training Setting}

Each model was trained for up to 300 epochs, with checkpoints selected based on the lowest validation loss. Training used a modified Mean Absolute Error (MAE) loss in which ground-truth GPP values were smoothed with Locally Weighted Scatterplot Smoothing (LOWESS) \citep{cleveland1979lowess} to reduce short-term noise and emphasize broader temporal trends.

\subsection{Model Evaluation}

Model evaluation focused on three aspects:

\subsubsection{Overall Performance}

Performance was assessed across four conditions: the full time series (GPP$^{\textrm{Overall}}$), the growing seasons only (GPP$^{\textrm{Growing}}$, May-September), negative extreme events (GPP$^{-}$), and positive anomalies (GPP$^{+}$). We used site-level Normalized Root Mean Squared Error (NRMSE) for comparability across ecosystems.

\subsubsection{Memory Retention}

Model memory retention was assessed through a permutation-based analysis. For each timestep $t$, we predicted $\mathrm{GPP}_t = f(\mathbf{x}_t, \mathbf{x}_{t-1}, \ldots, \mathbf{x}_{t-k+1})$ using a 120-day context window $k$. To evaluate the contribution of temporal information within this context window, the inputs corresponding to older timesteps than a given time lag $\tau$, i.e. ($x_{t-\tau-1},\ldots,x_{t-k+1}$), were permuted across samples. The time order of the most recent days ($\mathbf{x}_t,\ldots,\mathbf{x}_{t-\tau}$) remained intact. Decreasing $\tau$ therefore progressively shortened the effective context window, revealing how each model relied on distant versus recent observations. The resulting change in NRMSE quantified each model’s reliance on past information.

\subsubsection{Modality Importance}

An additional permutation analysis was applied at the modality level (S2, S1, LST, and R$_{so}$) to estimate relative modality contributions. For each modality, inputs were permuted independently while keeping others fixed, and the resulting increase in NRMSE relative to the unperturbed baseline indicated its Feature Importance (FI).

\section{Results and Discussion}

\subsection{Overall Performance}

\begin{figure}
	\centering
	\includegraphics[width=0.8\columnwidth]{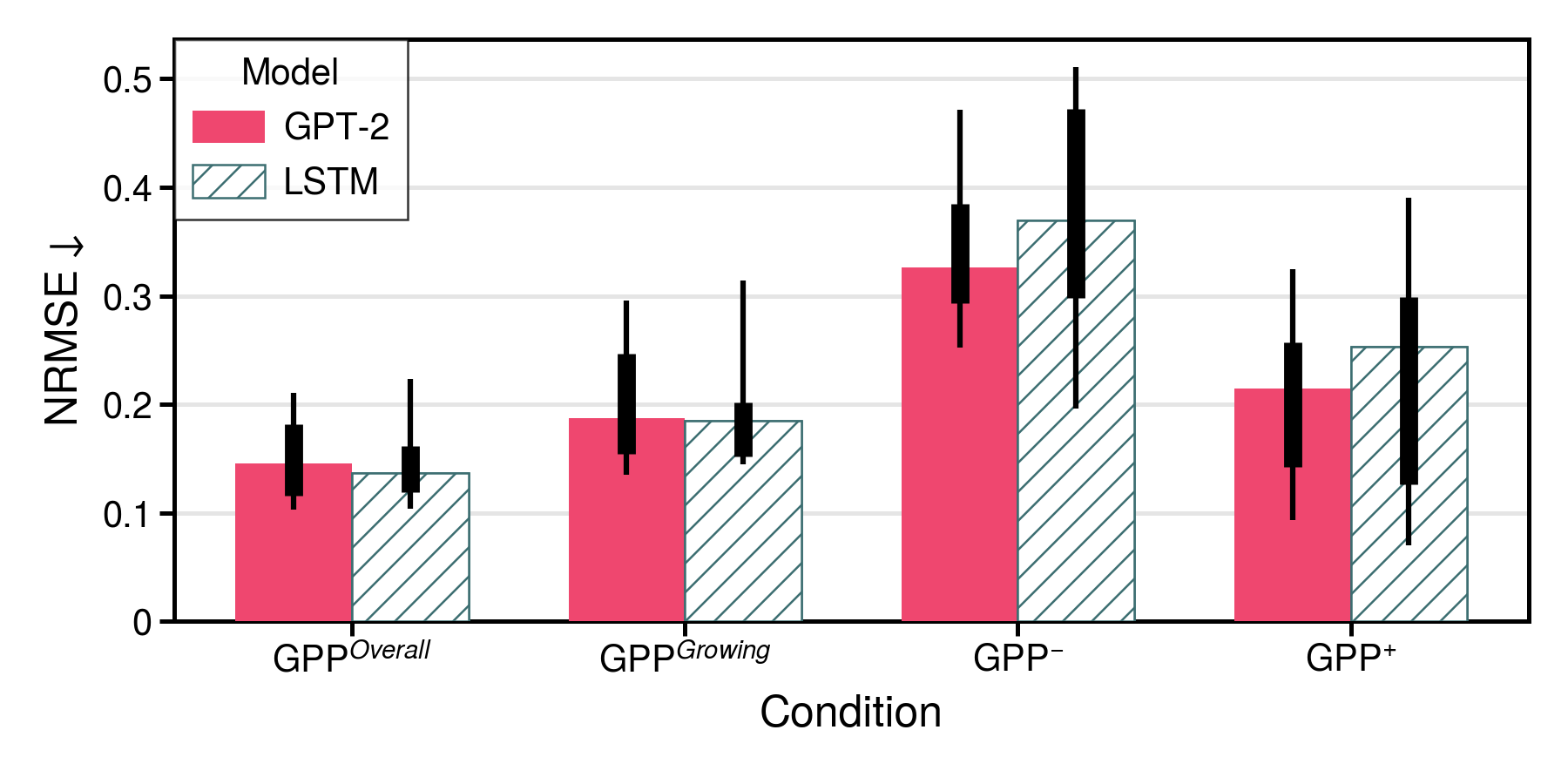}
        \caption{Normalized Root Mean Squared Error (NRMSE) for daily GPP predictions across four conditions: all timesteps (GPP$^{\textrm{Overall}}$), growing season (GPP$^{\textrm{Growing}}$), climate-induced extremes (GPP$^{-}$), and peaks of photosynthetic activity (GPP$^{+}$). Bars represent the median NRMSE, boxes the interquartile range (Q1-Q3), and whiskers span the 5th to 95th percentile across sites.}
	\label{fig:overall-performance}
\end{figure}

Both architectures accurately captured the seasonal dynamics of forest GPP (Fig.~\ref{fig:overall-performance}). Under typical conditions (GPP$^{\textrm{Overall}}$ and GPP$^{\textrm{Growing}}$), both models achieved low median NRMSE values, with LSTM performing slightly better (0.137 vs. 0.145 for GPP$^{\textrm{Overall}}$; 0.185 vs. 0.188 for GPP$^{\textrm{Growing}}$). These results indicate that both architectures effectively reproduced seasonal productivity patterns.

Model performance diverged during extreme events. In GPP$^{-}$ (stress-induced declines), GPT-2 achieved lower median errors (0.326 vs. 0.370 for LSTM) and tighter interquartile ranges, indicating higher robustness across sites. A similar trend was observed for GPP$^{+}$ (productivity peaks), with GPT-2 yielding lower errors (0.215 vs. 0.253). Both models showed larger uncertainties during GPP$^{-}$ than GPP$^{+}$, consistent with the stronger physiological decoupling from seasonality under drought and heat stress.

\subsection{Memory Retention}
\label{sec:results-memory-retention}

\begin{figure}
	\centering
	\includegraphics[width=0.8\columnwidth]{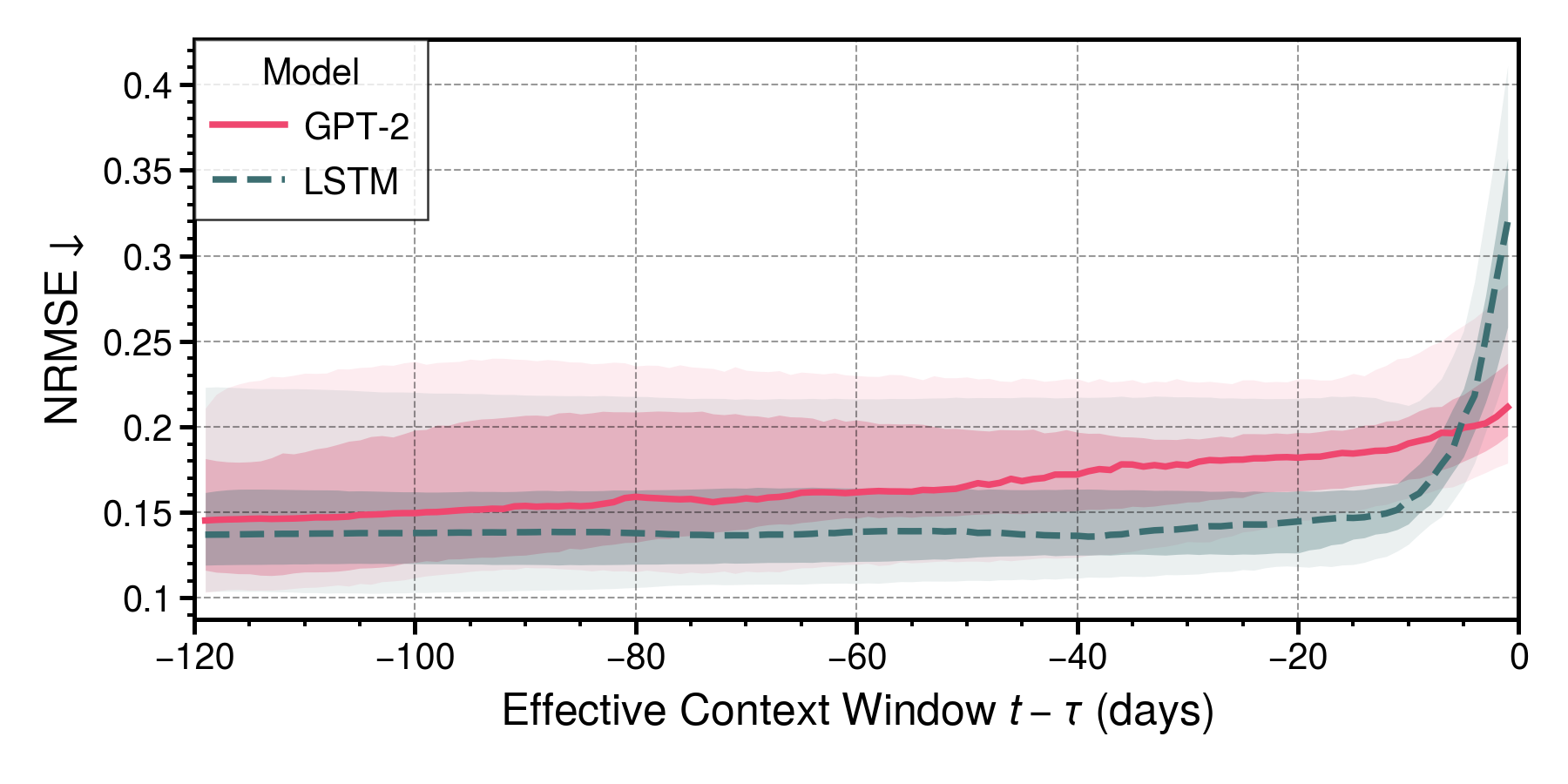}
        \caption{Permutation-based memory retention analysis across the 120-day context window. The y-axis shows the resulting NRMSE after prediction on the permuted inputs. Lines represent the median NRMSE across sites, while shaded areas show the interquartile range (darker) and the 5th-95th percentile range (lighter).}
	\label{fig:memory-retention}
\end{figure}

Both architectures exhibited stable performance across most of the sequence (Fig.~\ref{fig:memory-retention}), with NRMSE increasing only during shorter context windows. However, the wide variability across sites implies differing degrees of temporal dependence among forest types and conditions. This indicates that reliance on short-term information and limited use of long-term history is problematic and limits model performance.

\begin{figure}
	\centering
	\includegraphics[width=0.8\columnwidth]{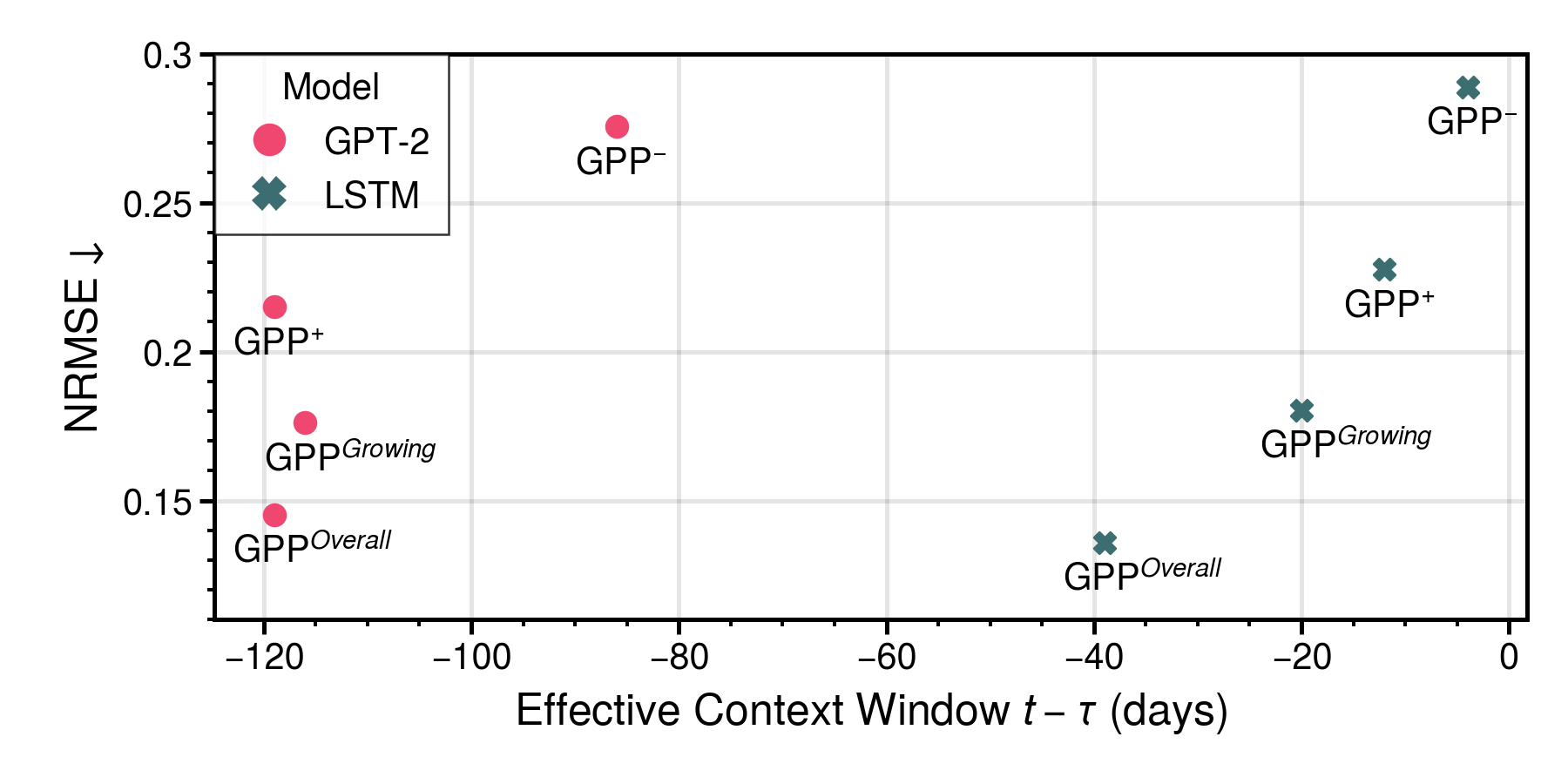}
        \caption{Minimum NRMSE achieved by each model for each GPP condition, plotted against the $t-\tau$ value in the 120-day context window where it was observed. Each point represents the condition-specific minimum NRMSE across the permuted sequence for either GPT-2 or LSTM.}
	\label{fig:memory-minima}
\end{figure}

The NRMSE increased more sharply with shorter context windows for the LSTM, suggesting a stronger dependence on near-past observations than GPT-2. The latter displayed smoother degradation, implying a more homogeneously distributed temporal attention. To further investigate these patterns, a minimum-error analysis was performed by identifying the context window at which each model achieved its lowest NRMSE (Fig.~\ref{fig:memory-minima}). GPT-2 consistently reached its minimum at deeper temporal horizons, while LSTM peaked much closer to the present. LSTM reached optimal performance with shorter sequences ($\approx$49 days for GPP$^{\textrm{Overall}}$, $\approx$20 days for GPP$^{\textrm{Growing}}$), while GPT-2 required longer temporal context ($\approx$119 and $\approx$116 days, respectively). During GPP$^{-}$ events, GPT-2 achieved its best performance (NRMSE = 0.275) at $\approx$86 days, whereas LSTM’s minimum (0.289) occurred at $\approx$4 days; similar behavior was found for GPP$^{+}$ events (0.215 at $\approx$119 vs. 0.227 at $\approx$12).

These results confirm that LSTM efficiently captures short-term dependencies, whereas GPT-2 more effectively integrates long-term dependencies, benefiting from its attention mechanism to leverage distant context. This is particularly relevant when lagged vegetation responses emerge during stress events, but it also reflects the model’s higher data demands, requiring longer input sequences to perform optimally. Overall, while both models effectively represent seasonal GPP variability, Transformers provide a distinct advantage in modelling GPP during climate extremes. However, the performance gap between models is modest: LSTM achieves comparable accuracy using substantially shorter context windows. This trade-off suggests that while GPT-2 can leverage longer sequences, its higher computational cost and need for larger training datasets may not always justify the marginal accuracy gains for typical GPP estimation tasks.

\subsection{Modality Importance}

\begin{figure}
	\centering
	\includegraphics[width=0.8\columnwidth]{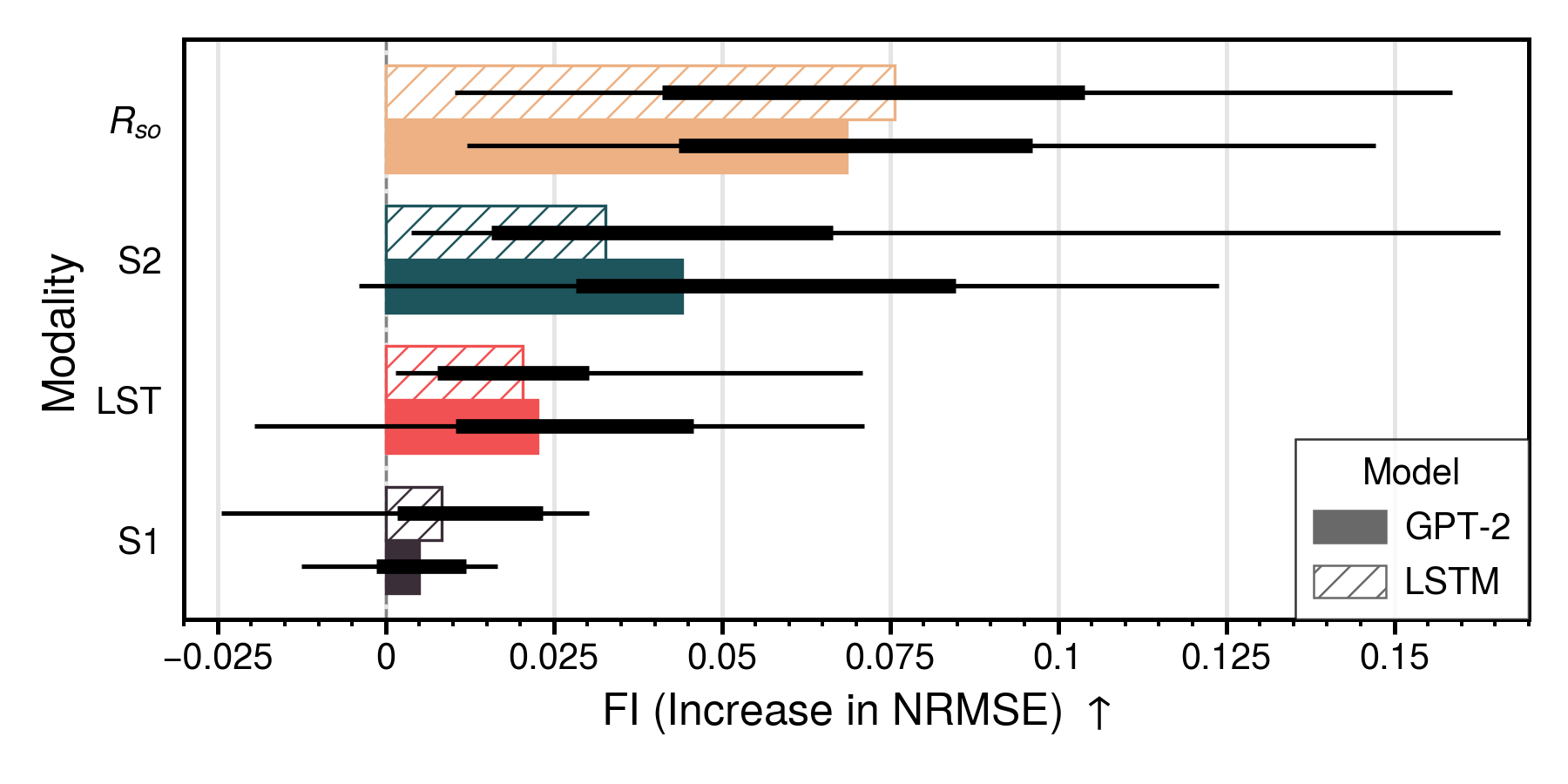}
    \caption{Feature importance (FI) values for each input modality. Bars represent the median FI across sites, with thick and thin error bars indicating interquartile and 5-95\% ranges, respectively.}
	\label{fig:feature-importance}
\end{figure}

Permutation-based feature importance (FI) analysis quantified the contribution of each input modality to model performance (Fig.~\ref{fig:feature-importance}). Both architectures exhibited consistent modality rankings: R$_{so}$ was the most informative feature, followed by S2, LST, and S1, which contributed minimally. Variability across sites was moderate, with R$_{so}$ and S2 showing the largest interquartile ranges, indicating spatial heterogeneity in their relevance.

R$_{so}$’s dominant importance highlights its effectiveness as a physically grounded spatiotemporal encoding. By integrating both solar geometry and site location, it conditions predictions across latitudinal and seasonal gradients, providing an ecophysiologically meaningful prior for photosynthetic activity. Unlike abstract temporal encodings, R$_{so}$ directly represents the solar energy available to drive GPP, explaining its consistently high FI values in both models.

S2 followed closely, reflecting the strong sensitivity of optical reflectance and VIs to canopy greenness and phenology. Its slightly lower and more variable importance relative to R$_{so}$ suggests that while the state of vegetation is essential for estimating GPP, it alone cannot capture the full temporal and physiological variability, particularly under stress conditions when greenness decouples from carbon uptake.

LST ranked third, although less influential overall. It likely provides complementary information during extremes by reflecting changes in canopy temperature and evaporative stress that optical indices cannot detect. This supports previous findings that thermal observations enhance detection of drought-related GPP reductions \citep{hoekvandijke2023comparing_forests_grasslands_drought,bayat2018integrating}.

S1 showed negligible importance, with FI values near zero for both models. This likely reflects the limited sensitivity of C-band backscatter to sub-canopy moisture, soil water content and structural variations relevant to GPP. Moreover, its higher noise and weaker physical linkage to photosynthetic processes may have discouraged the models from exploiting this modality.

\section{Conclusion}

This study compared a recurrent (LSTM) and a Transformer-based (GPT-2) architecture for predicting daily forest GPP from multimodal remote sensing inputs. Both models reproduced the seasonal dynamics of productivity, yet GPT-2 achieved higher accuracy and stability during climate-induced extremes, reflecting its greater ability to integrate information over extended temporal horizons. LSTM remained highly competitive under normal conditions and required substantially shorter context windows, highlighting an accuracy-efficiency trade-off between both architectures.

Among the input modalities, clear-sky solar radiation (R$_{so}$) was the most informative predictor, acting as a physically meaningful spatiotemporal encoding. Optical Sentinel-2 features followed closely, while LST contributed complementary thermal information during stress events. Sentinel-1, by contrast, showed little relevance.

These findings should be interpreted in light of several constraints. The 2016-2020 period and the restricted context window limited the number of extreme events and long-term responses captured. Extending the temporal record and incorporating additional modalities could better reveal how memory modulates GPP under normal conditions and extreme events. Such advances may turn predictive frameworks into early-warning systems for ecosystem stress and resilience monitoring.

\section*{Acknowledgment}
This study was financially supported via the “RS4BEF” project (iDiv’s Flexpool); and the “Digital Forest” project, Ministry of Lower-Saxony for Science and Culture (MWK) via the program Niedersächsisches Vorab (ZN 3679).

\bibliographystyle{abbrvnat}
\bibliography{references}

\end{document}